# Towards Automatic Recognition of Pure & Mixed Stones using Intraoperative Endoscopic Digital Images

**Authors:**

Vincent Estrade[1], Michel Daudon[2], Emmanuel Richard[3], Jean-Christophe Bernhard[1], Franck Bladou[1], Grégoire Robert[1], Baudouin Denis de Senneville[4]

[1]Department of Urology, CHU Pellegrin, Place Amélie Raba Léon 33000 Bordeaux, France

[2]Department of Multidisciplinary Functional Explorations, AP-HP, Tenon Hospital, INSERM UMRS 1155, Sorbonne University, Paris, France

[3]University of Bordeaux, INSERM, BMGIC, U1035, CHU Bordeaux, 33076 Bordeaux, France

[4]University of Bordeaux, IMB, UMR CNRS 5251, 351 Cours de la Libération, F-33405 Talence Cedex, France

Vincent Estrade (vincent.estrade@gmail.com)

Michel Daudon (michel.daudon@aphp.fr)

Emmanuel Richard (emmanuel.richard@chu-bordeaux.fr)

Jean-Christophe Bernhard (jean-christophe.bernhard@chu-bordeaux.fr)

Franck Bladou (franck.bladou@chu-bordeaux.fr)

Grégoire Robert (gregoire.robert@chu-bordeaux.fr)

Baudouin Denis de Senneville (bdenisde@math.u-bordeaux.fr)

**Abbreviations:**

ESR: Endoscopic Stone Recognition

LASER: Light Amplification by Stimulated Emission of Radiation

CT: Computed Tomography

IR: InfraRed

FTIR: Fourier Transform InfraRed spectroscopy

CCD: Charge-Coupled Device

COM: Calcium Oxalate Monohydrate

COD: Calcium Oxalate Dihydrate

UA: Uric Acid

AI: Artificial Intelligence

CNN: Convolutional Neural Network

Grad-CAM: Gradient-weighted Class Activation Mapping

VE: Vincent Estrade

DSDECT: Dual-Source Dual-Energy Computed Tomography

AUROC: Area-Under-the-ROC curve

PPV: Positive predictive value

NPV: Negative Predictive Value

FPR: False Predictive Rate

FNR: False Negative Rate

TFL: Thulium Fiber Laser

**ABSTRACT**

**Objective:** To assess automatic computer-aided *in-situ* recognition of morphological features of pure and mixed urinary stones using intraoperative digital endoscopic images acquired in a clinical setting.

**Materials and methods:** In this single-centre study, a 20-year experienced urologist intraoperatively and prospectively examined the surface and section of all kidney stones encountered. Calcium oxalate monohydrate (COM/Ia), dihydrate (COD/IIb) and uric acid (UA/IIIb) morphological criteria were collected and classified to generate annotated datasets. A deep convolutional neural network (CNN) was trained to predict the composition of both pure and mixed stones. To explain the predictions of the deep neural network model, coarse localisation heat-maps were plotted to pinpoint key areas identified by the network.

**Results:** This study included 347 and 236 observations of stone surface and stone section, respectively; of all stones, around 80 % exhibited only one morphological type and around 20 % displayed two. A highest sensitivity of 98 % was obtained for the type "pure IIIb/UA" using surface images. The most frequently encountered morphology was that of the type "pure Ia/COM"; it was correctly predicted in 91 % and 94 % of cases using surface and section images, respectively. Of the mixed type "Ia/COM+IIb/COD", Ia/COM was predicted in 84 % of cases using surface images, IIb/COD in 70 % of cases, and both in 65 % of cases. Concerning mixed Ia/COM+IIIb/UA stones, Ia/COM was predicted in 91 % of cases using section images, IIIb/UA in 69 % of cases, and both in 74 % of cases.

**Conclusions:** This preliminary study demonstrates that deep convolutional neural networks are promising to identify kidney stone composition from endoscopic images acquired intraoperatively. Both pure and mixed stone composition could be discriminated. Collected in a clinical setting, surface and section images analysed by deep CNN provide valuable information about stone morphology for computer-aided diagnosis.

# INTRODUCTION

Modern endoscopic treatment of urinary stones now relies on LASER (Holmium-Yag) fragmentation of stones, which can be performed using "popcorn" [1], "dusting" modes [2], or more recently by means of Thulium Fiber LASER (TFL) [3-4]. However, LASER fragmentation may destroy the morphology of the targeted stone [5]. Yet, analysis of stone morphology is crucial for an aetiological diagnosis of stone disease [6-8] and for the development of novel immediate post-operative treatment strategies that will eliminate potential residual stone fragments with a lower probability of relapse [9]. For example, calcium oxalate monohydrate (COM/Ia) or dihydrate (COD/IIb) criteria would support the prescription of an immediate diet with potassium citrate, as reported in [10]. Recognition of uric acid (UA/IIIb) morphological criteria would steer the medical decision towards a postoperative urinary alkalinization with potassium citrate or sodium bicarbonate to dissolve residual fragments [11].

The complete morphological analysis workflow may typically include the following two complementary steps:

1. An intraoperative step, which is conducted by a urologist, involves an endoscopy-based examination of the morphology of entire stones *in-situ* before their destruction. This step is commonly referred to as endoscopic stone recognition (ESR) [12]. Endoscopic images can be conveniently obtained before (surface image) and after (section images) fragmentation, thus providing valuable morphological information. Estrade et al. recently showed that ESR allowed identifying the following morphologies: calcium oxalate monohydrate (COM), also referred to as the types Ia, Ib, Id or Ie (subscripts in the Latin alphabet differentiate morphological subtypes, each being associated with a specific aetiology), calcium oxalate dihydrate (COD, or IIa/IIb), uric acid (UA, or IIIa/IIIb), carbapatite (or IVa1), carbapatite and struvite (or IVb), brushite (or IVd), and cystine (or Va) [12].

2. A post-operative step, which is performed by a biologist, consists of collecting morpho-constitutional stone information based on both microscopic morphological, *i.e.*, binocular magnifying glass, and spectrophotometric infrared recognition (FTIR analysis) [6-8].

The international morpho-constitutional classification of urinary stones includes seven groups (denoted by roman numerals "I" to "VII"), each being associated with a specific crystalline type (I = whewellite, II = weddellite, III = uric acid and urates, IV = calcium and non-calcium phosphates, and V = cystine. groups VI and VII are devoted to other stones). Each group comprises several subgroups that differentiate morphologies and aetiologies for a given crystalline type. Furthermore, urinary stones have mixed morphologies, *i.e.*, include at least two morphologies (almost half of all cases are concerned). The interested reader is referred to [13] for additional information about the international morpho-constitutional classification of urinary stones.

Recently, an artificial intelligence (AI) algorithm applied to various types of microscopic images of stones *ex-vivo* proved to be a promising asset for automatic ESR using both peri- and post-operative images. While Serrat et al. fed texture and colour features of stones into a random forest classifier [14], Black et al. obtained much improved scores using a deep convolutional neural network (CNN) [15]. However, both approaches used *ex-vivo* stone fragments placed in a controlled environment. Images were not disturbed by motion blur, specular reflections or scene illumination variations, as occurs in common practice during an intraoperative endoscopic imaging session. More recent works demonstrated the potential of automated ESR approaches using *in-vivo* images acquired in clinical conditions with ureteroscopes on three types of pure stones - Ia/COM, IIb/COD, and IIIb/UA - from 125 kidney stone images [16][17].

However, morphological examinations of the entire stone before its destruction provide the best diagnostic agreement [6-8,12,13]. Moreover, Corrales et al. showed that almost half of all urinary stones are of mixed morphologies with two or even three different crystalline components [13]. AI applications must therefore be improved to meet this challenge. The present study has three objectives:

1. To report the preliminary results of the automatic ESR of the morphological components of both pure and mixed urinary stones (these mixed stones being composed of two morphologies in the scope of this study), *in-situ*, and using intraoperative endoscopic images acquired in a clinical setting. Thus, the images used in this project were captured in an uncontrolled environment by means of ureteroscopes. Besides, the overall performance of a deep neural network was assessed in this setting.

2. To analyse diagnostic scores computed from images obtained before and after LASER fragmentation.

3. To attempt to understand decisions made by a deep neural network in this setting. Specifically, a common problem of deep CNNs is their inability to explicitly display what the model has learned, hence often their naming "black box" algorithms. Predictions computed from deep CNNs are in turn hard to explain. Currently there is a growing interest in the development of robust validation procedures to address this key issue. In this study, we illustrate the usefulness of providing deep CNN-algorithm based "attention" maps to understand where the algorithm was "looking" in the endoscopic image when it took its decision.

**MATERIALS & METHODS**

**Study design**

A urologist (VE, 20 years of experience) prospectively examined intraoperative endoscopic digital images of stones acquired between January 2018 and November 2020 in a single centre using a flexible digital ureterorenoscope (Olympus URF-V CCD sensor). First the endoscopic examination included a visual observation of the stone surface. Then, a LASER-induced stone split in two parts was performed (LASER (Holmium-Yag) parameters: frequency = 5 Hz, energy = 1.2–1.4 J, power = 6–7 W, pulse length = short, fibre diameter = 230 or 270 μm). A second visual observation of the section was then performed. An additional fragmentation session was carried out when needed, thus allowing fragmentation of all types of pure and mixed stones. Subsequently, ESR was confirmed by means of microscopic observations of LASER-fragmented stones based on both morphological, *i.e.*, binocular magnifying glass, and infrared, *i.e.*, FTIR, analyses. The study adhered to all local regulations and data protection agency recommendations (National Commission on Data Privacy requirements). Patients were informed that their data would be used anonymously.

**Morphological criteria**

Morphological criteria were collected and classified according to recommendations outlined in [12]. Stones composed of Ia/COM, IIb/COD and IIIb/UA morphologies were selected. Hence, five morphology classes were included in this study, with three pure stones (Ia/COM, IIb/COD and IIIb/UA) and two mixed stones divided into two morphologies (Ia/COM + IIb/COD and Ia/COM + IIIb/UA).

**Computer-assisted ESR analysis**

*Generation of annotated datasets*

Two annotated datasets were generated: the first comprised surface images (referred to as "surface dataset" throughout the rest of the manuscript), the second contained section images (referred to as "section dataset"). All images were automatically cropped and resampled to an equal size of 256-by-256 pixels, then served as input of the automatic ESR algorithm.

*Automatic ESR algorithm*

A deep CNN was trained to predict the composition of both pure and mixed stones. Used as a multi-class classification model, the deep CNN was a ResNet-152-V2 [18]. The optimizer algorithm for training the deep learning model was Adam (learning_rate=0.001) [19]. The loss function was a categorical cross-entropy (). The batch size was 8 and 100 epochs were performed. To improve the ability for the network to generalize, the training dataset was expanded through data augmentation.

In our implementation, horizontal/vertical flips and affine transformations, including random combinations of scaling (range=0.3), rotation (range=50°), and translation (range=0.2 of total width/height) were applied during training.

Two networks were built separately: one using the "surface dataset" and the other using the "section dataset".

*Activation maps*

To explain the predictions of the deep CNN, coarse localization heat-maps were plotted to pinpoint key areas identified by the network. To this end, activation maps using the "Gradient-weighted Class Activation Mapping" method (also referred to as "Grad-CAM") were displayed, as discussed by Selvaraju and colleagues [20].

*Implementation details*

Our implementation was performed using TensorFlow 1.4 and Keras 2.2.4. The Keras image preprocessing tools available at https://keras.io/api/preprocessing/image/ were applied for data augmentation.

**Statistical analysis**

*Quantitative assessment of automatic ESR*

For both "surface dataset" and "section dataset", stones were randomly divided into complementary training (70 %) and testing (30 %) subsets (stratified split/no redundancy). A cross-validation step was repeated 10 times with randomly shuffled combinations for training and testing. The full process was also repeated with different random initialization seeds for the deep CNN algorithm. Average test metrics were reported for each step: accuracy, area under the ROC curve (AUROC), specificity, sensitivity, positive predictive value (PPV), negative predictive value (NPV), false predictive rate (FPRs) and false negative rate (FNRs). For additional information about these test metrics, see [21] [22]. Concerning mixed stones, test metrics were evaluated when at least one of the pure morphologies was predicted (we recall that mixed stones are composed of two pure morphologies in the scope of this study) and when both morphologies were predicted.

*Qualitative assessment of activation maps*

A qualitative (visual) observation of the activation maps was carried out for surface and section images individually. The amount of correctly classified and misclassified images was calculated when the hot spots in the activation maps were located: a) in the stone, b) outside the stone, c) over the tip of the endoscope.

## RESULTS

**Stone characteristics**

The study included 347 observations of stone surface (pure stones: Ia = 191/150 [number of images/number of unique stones], IIb = 53/48, IIIb = 29/23; mixed stones: Ia + IIb = 64/54, Ia + IIIb = 10/9) and 236 observations of stone section (pure stones: Ia = 127/96, IIb = 30/29, IIIb = 25/22; mixed stones: Ia + IIb = 31/26, Ia + IIIb = 23/15).

Figure 1 shows representative examples of *in-situ* endoscopic images obtained for each pure stone morphology before LASER fragmentation (surface image). Images acquired after LASER fragmentation (section images) are shown in Figure 2. The three pure stone morphologies (first three rows of figures 1 and 2) had the following visual characteristics:

- **Ia/COM:** before LASER fragmentation (1a): a smooth or mammillary, dark-brown surface ; after LASER fragmentation (2a): compact concentric layers with a radiating organization starting from a nucleus.
- **IIb/COD:** before LASER fragmentation (1c): a yellowish or light-brown surface with smooth, long bi-pyramidal crystals (like small desert roses) ; after LASER fragmentation (2c): a compact poorly organized pale brown-yellow crystalline section.
- **IIIb/UA:** before LASER fragmentation (1e): a rough, porous surface with heterogeneous, beige to orange-red colour; after LASER fragmentation (2e): poorly organized, porous ochre to orange structure.

Representative examples of *in-situ* endoscopic images of Ia + IIb and Ia + IIIb mixed stones are also shown (figure 1 and 2, last two rows). For each, the combinations of the corresponding two of the three pure morphologies mentioned above are visible.

**Diagnostic performance of automatic ESR**

The testing subset of the "surface dataset" included 105 urinary stones (pure stones: Ia = 57, IIb = 16, IIIb = 9; mixed stones: Ia + IIb = 20, Ia + IIIb = 3). The testing subset of the "section dataset" included 70 urinary stones (pure stones: Ia = 38, IIb = 9, IIIb = 7; mixed stones: Ia + IIb = 9, Ia + IIIb = 7).

Table 1 details the diagnostic performance of the deep CNN classifier for each tested pure type. The best sensitivity was obtained for the type IIIb using surface images (98 % of IIIb stones correctly predicted). The most frequently encountered morphology was the type "pure Ia"; it was correctly predicted in 91 % and 94 % of cases using surface and section images, respectively. On average, the accuracy was higher than 87 % for both pure and mixed stones.

Table 2 details the diagnostic performance of the deep CNN classifier for each tested mixed type. Concerning Ia + IIb stones, Ia was predicted in 84 % of cases using surface images, IIb in 70 % of cases, and both types in 65 % of cases. Concerning Ia + IIIb stones, Ia was predicted in 91 % of cases using section images, IIIb in 69 % of cases, and both types in 74 % of cases. These findings are also displayed in the confusion matrices shown in figure 3: at least one of the two morphologies constituting mixed stones is preferably detected as a secondary choice, as indicated by off-diagonal values in figure 3. Overall, percentages of valid predictions using surface and section images were equal to 83 % and 81 %, respectively, see blue cells in confusion matrices in figure 3.

**Qualitative performance of activation maps**

Figure 1 and 2 also show image areas where the classification network concentrated attention in surface (figure 1) and section datasets (figure 2), respectively. Activation maps were overlaid on the digital endoscopic image in order to establish whether the classification model relied on relevant urological regions in the decision-making process. For example, a hot spot was usually observed in (1a) on a mammillary dark-brown area, which is a hallmark of Ia. Hot spots were found on characteristic stone features in 98 % of the correctly classified images (using either surface or section images, see "True positive" columns in figure 1 and 2). Hot spots outside the stone were found in 33 % and 25 % of misclassified surface and section images, respectively (figure 1d). Similar the tip of the endoscope was present in the image field of view in 5 % and 2 % of misclassified surface and section images, respectively (red arrow in figure 1h).

**DISCUSSION**

In this study, we evaluated a deep learning model to predict *in-situ* the morphology of pure and mixed stones based on intraoperative endoscopic digital images acquired in a clinical setting. A learning curve is needed to acquire the ESR skill which may limit its translation to practical use, especially when mixed stones morphologies are involved [12]. A computer-assisted approach can deliver reproducible results and minimises operator dependency while assisting visual interpretation of stone morphologies.

As reported in [8] and [12], ESR may be beneficial before fragmentation to preserve an etiological approach in lithiasis. The motivation is two-fold:

1. LASER fragmentation (Holmium-Yag and TFL), whether performed with "popcorn" [1] or "dusting" modes [2], irreversibly destroys the stone morphology. One can notice that post-operative FTIR examinations of the stone powder itself may not provide sufficient information for the lithogenic stage [6-9].

2. The IR spectra can be modified when the stone fragmentation is achieved in dusting mode with high frequency TFL [2-5]. This may bias in turn FTIR dust examinations: one can observe IR changes from COD towards COM, IR changes towards an amorphous phase in carbapatite, IR changes towards a differing and amorphous crystalline phase in Magnesium Ammonium Phosphate and IR changes from brushite towards carbapatite [5].

We focused on pure and mixed stones involving Ia/COM, IIb/COD and IIIb/UA morphologies. This strategy was driven by the epidemiological distribution of occurrence of urinary stones to obtain a sufficiently large population for an opposable statistical approach [23]. These morphological types cover almost 85% of the most common stones that urologists encounter on a daily practice [6]. Our datasets of endoscopic images will be supplemented by more pure and mixed stone images in future studies (ongoing at our institution) in order to increase the number of morphologies to be predicted. In addition, the automatic ESR score will likely improve if the network is able to train on a larger set of data.

It must be reported that particles flying around in the saline may disturb the morphological stone examination. To properly recognize the colors and textures of the stone surface, it is necessary to wait a few tens of seconds until the saline solution has cleaned the urine in the kidney cavities. Then, once the stone is spitted into two parts, a few seconds are again mandatory before the saline solution has cleaned micro-particles of the stone. In the current study, a ureteral access sheath was used to improve saline flow. However, in practice, in the absence of a ureteral access sheath, only a few additional seconds are needed for a complete cleaning using the saline serum. It must be underlined that particles flying around in the saline were not present in the images used for training in the current study.

It must also be underlined that a sufficient stability of the endoscopic video image is mandatory for a short duration (5 to 10 seconds) to obtain good still frames. Any motion event is likely to hamper the image quality and to bias, in turn, the predictions of our trained network. In the current work, the trained urologist (VE) made several attempts to get sharp screenshot images (2 attempts in average, max = 4). Several strategies may be investigated in future works to improve the performance of the method on motion-corrupted endoscopic images. Enhanced high quality images may be obtained from low quality image series using dedicated motion compensated super resolution techniques [24]. In addition, data augmentation techniques involving simulated blur and motion events may further improve the ability of the network to generalize for motion corrupted endoscopic images [25].

Any unobserved events/image artifacts during the training step may disturb in turn the predictions of our trained network. In future studies, automated and reliable quality control on the input images must be developed in order to detect potential failure modes of the network. The urologist will then be advised to take LASER-fragmented stones for a post-operative infrared - FTIR - examination in a dedicated laboratory.

Processing of both surface and section images provide valuable information about stone morphology for computer-aided diagnosis. In practice, recognition of the morphological hallmarks in surface images is easier than that in section images (or even at the nucleus), as shown in [12] and [26]. Consequently, our surface dataset was generally more densely populated than our section one (except for mixed Ia+IIIb stones for which section images better reveal the two morphological types). However, the diagnostic performance of the surface dataset was found to be comparable to that obtained in the section dataset (figure 3, blue cells). On the other hand, the diagnostic performance of the section dataset was better than that obtained with the surface dataset for pure IIIb/UA (see Table 1). Surface and section images may thus provide a source of cross-validation of the diagnosis according to both complementary and redundant information. We believe that paired surface and section images for each stone may be incorporated into the CNN in order to improve the accuracy of the predictions.

The annotated datasets must be accurate since any subjectivity in ESR or potential bias of the urologist may be transferred into the network model. A concordance study between endoscopic digital pictures and microscopy may provide confirmed ESR image of stones corresponding to specific aetiologies or lithogenic mechanisms [12]. In addition to automatic ESR, activation maps could become an important tool to test whether, during the decision-making process, the classification model relied on relevant urological regions. Moreover, our study indicated that a hot spot located outside the stone led to a misclassification.

Deep CNNs capable of processing a large number of specific images efficiently are paving the way for automatic ESR on videos, which would further improve the accuracy of classification scores. This will require the development of dedicated algorithms to remove on-the-fly irrelevant areas of the image that are likely to bias the network, such as those around the endoscope tip and surrounding tissue, among others.

**CONCLUSION**

Combined with endoscopic digital images confirmed according to the criteria published in [12], AI is a good candidate for the automatic ESR of the morphological features of pure and mixed urinary stones composed of two morphologies. This study is a preliminary step towards the automatic ESR of mixed stones of several morphologies. Activation maps may prove to be a great asset for urologists to intraoperatively understand the predictions made by the AI model. This is especially crucial in medical applications where model accuracy is paramount. Combined with didactic boards of confirmed endoscopic images, both computer-aided diagnosis and associated activation maps may be useful for urologists to recognise stones *in-situ* using an endoscopic examination before destruction. The combination of automatic intraoperative ESR and post-operative infrared - FTIR - examinations of LASER-fragmented stones would improve the aetiological approach to lithiasis.

## ACKNOWLEDGMENT

Experiments presented in this paper were carried out using the PlaFRIM experimental testbed, supported by Inria, CNRS (LABRI and IMB), Université de Bordeaux, Bordeaux INP and Conseil Régional d'Aquitaine (see https://www.plafrim.fr/). The authors gratefully acknowledge the support of NVIDIA Corporation with their donation of a TITAN X GPU used in this research.

# LIST OF TABLES

**Table 1.** Diagnostic performance of implemented deep CNN classifier for pure stones (i.e., Ia/COM, IIb/COD and IIIb/UA morphologies). Results obtained using surface and section images are reported after cross-validation (averaged indicators shown with standard deviations). AUROC: Area-Under-the-ROC (Receiver Operating Characteristic) curve; PPV: positive predictive value; NPV: negative predictive value; FPR: false predictive rate; FNR: false negative rate. Accuracies, sensitivities, specificities, PPVs, and NPVs shown in percentages.

| Stone type | Accuracy (%) | AUROC | Sensitivity (%) | Specificity (%) | PPV (%) | NPV (%) | FPR (%) | FNR (%) |
|---|---|---|---|---|---|---|---|---|
| Surface | | | | | | | | |
| Ia | 90 ± 3 | 0.90 ± 0.03 | 91 ± 5 | 90 ± 4 | 92 ± 3 | 90 ± 4 | 10 ± 4 | 9 ± 5 |
| IIb | 93 ± 2 | 0.86 ± 0.04 | 77 ± 7 | 95 ± 2 | 76 ± 9 | 96 ± 1 | 5 ± 2 | 23 ± 7 |
| IIIb | 99 ± 1 | 0.98 ± 0.02 | 98 ± 5 | 99 ± 1 | 90 ± 8 | 100 ± 0 | 1 ± 1 | 2 ± 5 |
| Section | | | | | | | | |
| Ia | 94 ± 2 | 0.94 ± 0.02 | 94 ± 2 | 93 ± 5 | 94 ± 4 | 94 ± 3 | 7 ± 5 | 6 ± 2 |
| IIb | 94 ± 3 | 0.83 ± 0.09 | 69 ± 18 | 97 ± 2 | 77 ± 13 | 96 ± 3 | 3 ± 2 | 31 ± 18 |
| IIIb | 95 ± 2 | 0.78 ± 0.14 | 60 ± 30 | 97 ± 2 | 63 ± 27 | 97 ± 1 | 3 ± 2 | 40 ± 30 |

**Table 2.** Diagnostic performance of implemented deep CNN classifier for mixed stones (i.e., Ia/COM + IIb/COD and Ia/COM + IIIb/UA morphologies). Test metrics evaluated using surface and section images when at least one pure morphology is predicted (N.B. mixed stones composed of two pure morphologies in this study) and when both morphologies were predicted.

| Stone type | Predicted kidney type | Accuracy (%) | AUROC | Sensitivity (%) | Specificity (%) | PPV (%) | NPV (%) | FPR (%) | FNR (%) |
|---|---|---|---|---|---|---|---|---|---|
| | | | | Surface | | | | | |
| Ia+IIb | at least Ia | 89 ± 2 | 0.88 ± 0.03 | 84 ± 4 | 91 ± 2 | 85 ± 4 | 91 ± 2 | 9 ± 2 | 16 ± 4 |
| | at least IIb | 90 ± 2 | 0.82 ± 0.03 | 70 ± 6 | 94 ± 2 | 70 ± 6 | 94 ± 1 | 6 ± 2 | 30 ± 6 |
| | both Ia and IIb | 87 ± 3 | 0.78 ± 0.05 | 65 ± 10 | 92 ± 3 | 65 ± 8 | 92 ± 2 | 8 ± 3 | 35 ± 10 |
| Ia+IIIb | at least Ia | 94 ± 1 | 0.93 ± 0.02 | 89 ± 4 | 96 ± 1 | 91 ± 3 | 96 ± 2 | 4 ± 1 | 11 ± 4 |
| | at least IIIb | 98 ± 0 | 0.93 ± 0.04 | 86 ± 8 | 99 ± 0 | 87 ± 6 | 99 ± 0 | 1 ± 0 | 14 ± 8 |
| | both Ia and IIIb | 98 ± 1 | 0.75 ± 0.14 | 50 ± 28 | 100 ± 1 | 71 ± 32 | 99 ± 1 | 0 ± 1 | 50 ± 28 |
| | | | | Section | | | | | |
| Ia+IIb | at least Ia | 91 ± 2 | 0.90 ± 0.02 | 86 ± 4 | 93 ± 2 | 86 ± 4 | 93 ± 2 | 7 ± 2 | 14 ± 4 |
| | at least IIb | 91 ± 2 | 0.78 ± 0.06 | 60 ± 13 | 95 ± 1 | 64 ± 8 | 94 ± 2 | 5 ± 1 | 40 ± 13 |
| | both Ia and IIb | 88 ± 2 | 0.72 ± 0.08 | 51 ± 17 | 93 ± 2 | 51 ± 10 | 93 ± 2 | 7 ± 2 | 49 ± 17 |
| Ia+IIIb | at least Ia | 94 ± 2 | 0.93 ± 0.02 | 91 ± 3 | 95 ± 2 | 90 ± 4 | 96 ± 1 | 5 ± 2 | 9 ± 3 |
| | at least IIIb | 95 ± 2 | 0.83 ± 0.09 | 69 ± 18 | 97 ± 2 | 73 ± 13 | 97 ± 1 | 3 ± 2 | 31 ± 18 |
| | both Ia and IIIb | 94 ± 2 | 0.85 ± 0.09 | 74 ± 18 | 97 ± 2 | 74 ± 16 | 97 ± 2 | 3 ± 2 | 26 ± 18 |

# LIST OF FIGURES

**Figure 1.** Representative automatic ESR results obtained before LASER fragmentation (surface image). Examples of both correctly (left panel) and misclassified images (right panel; type reported on far left is not recognised by network) are shown. *In-situ* surface images (left image of each panel) are reported for each stone composition. Ia/COM, IIb/COD and IIIb/UA pure morphologies are reported in first three rows. For each mixed stone (last two rows), a mixture of the corresponding pure morphologies is visible. Activation maps (right image of each panel) show areas where network concentrates attention.

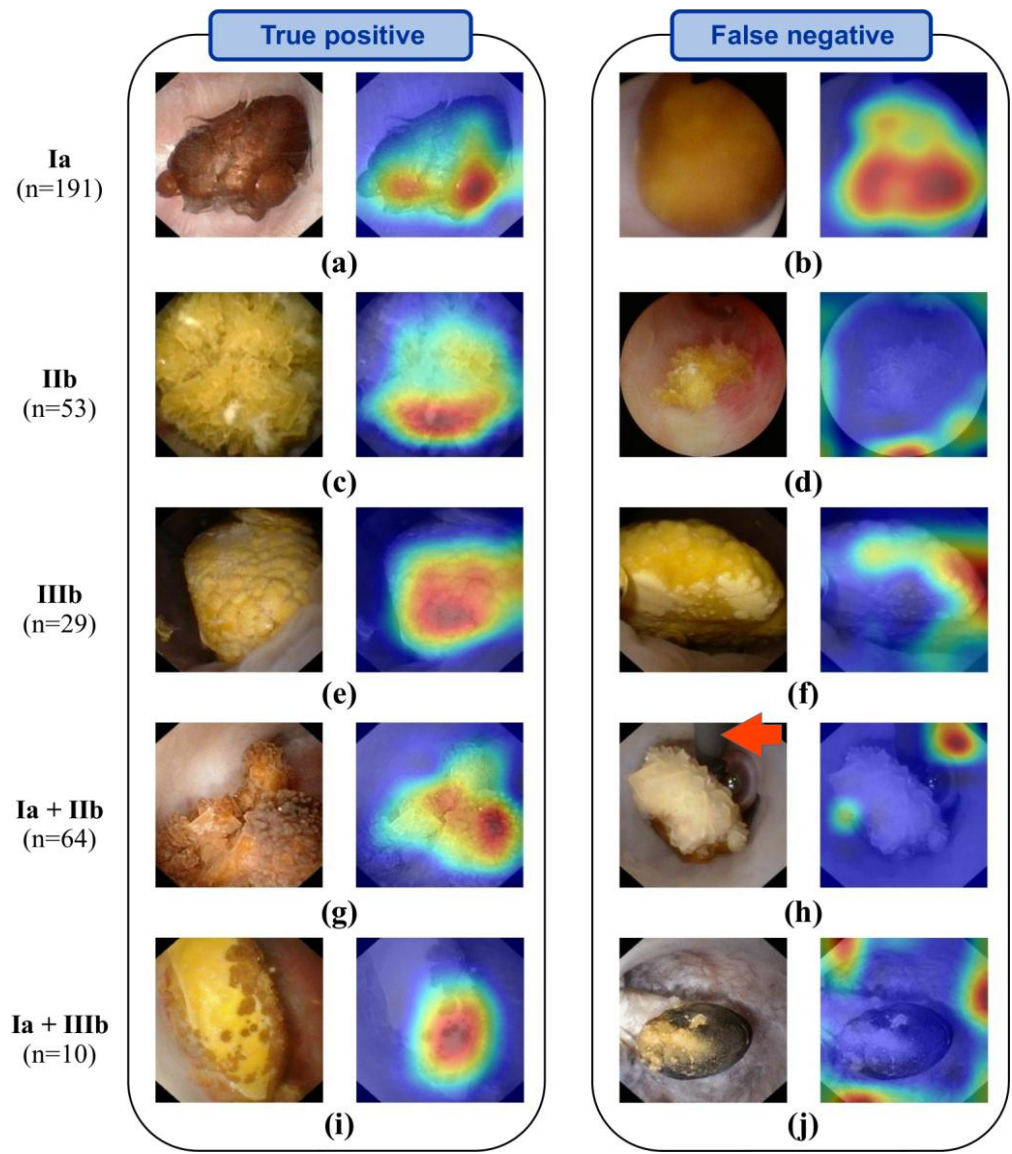

**Figure 2.** Representative automatic ESR results obtained after LASER fragmentation (section images). Examples of both correctly (left panel) and misclassified images (right panel: type reported on far left is not recognised by network) are shown. *In-situ* section images (left image of each panel) are reported for each stone composition. Ia/COM, IIb/COD and IIIb/UA pure morphologies are reported in first three rows. For each mixed stone (last two rows), a mixture of the corresponding pure morphologies is visible. Activation maps (right image of each panel) show areas where network concentrates attention.

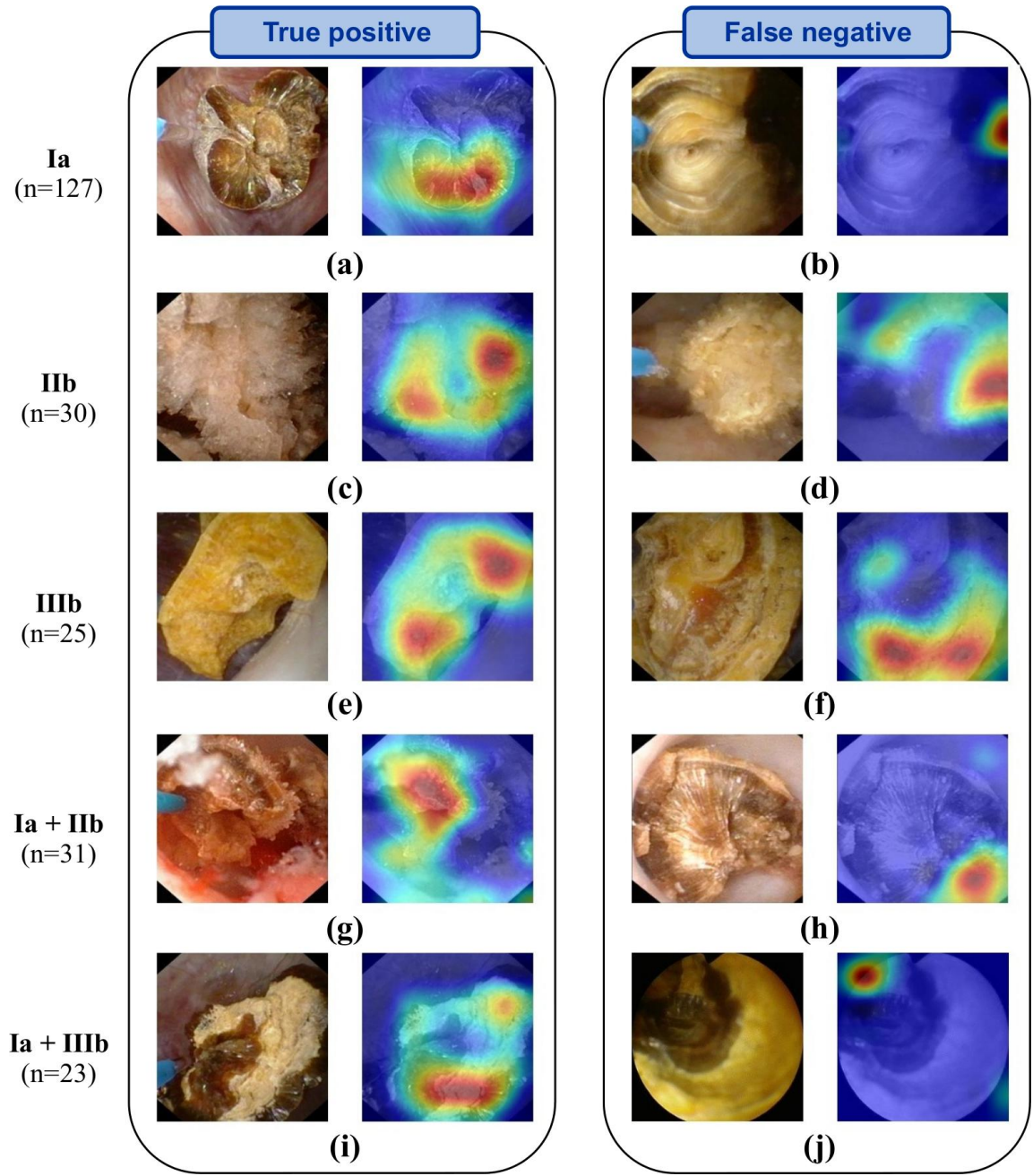

**Figure 3.** Confusion matrices for implemented deep CNN classifier obtained using surface (a) and section (b) datasets. Each column of the matrices represents an actual stone type while each line represents a predicted type. Green diagonal cells show number (averaged by cross-validation) and percentage of correct predictions by trained network. Red off-diagonal cells correspond to wrongly predicted observations. Column on far right shows PPV (green numbers) and false discovery rate (red numbers). Bottom row shows sensitivity (green numbers) and the false negative rate (red numbers). Blue cell bottom right shows overall percentage of correct (green) and incorrect (red) predictions.